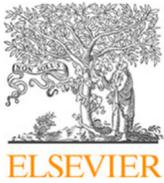
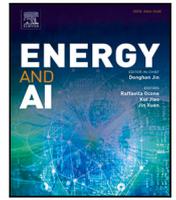

# Explaining deep neural network models for electricity price forecasting with XAI

Antoine Pesenti ⃝*, Aidan O'Sullivan

*UCL Energy Institute, University College London, United Kingdom*

## HIGHLIGHTS

- Five electricity markets are explained with XAI methods applied to their DNN models
- New SSHAP explainable method is introduced to address EPF models' high dimension
- XAI explanations reveal insights into the feature transformations inside the models

## GRAPHICAL ABSTRACT

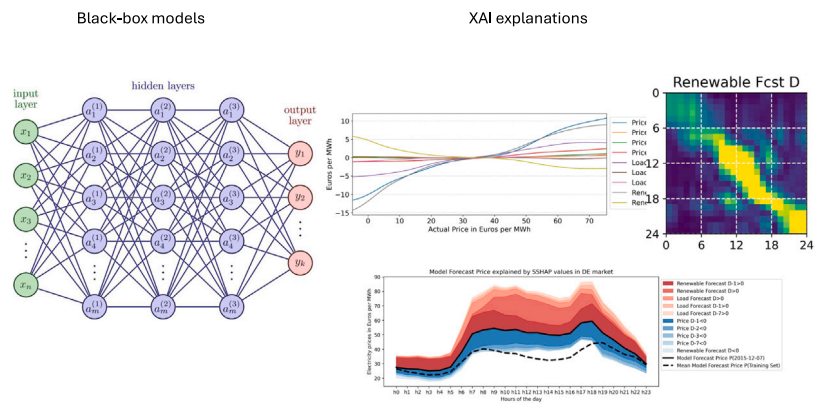

## ARTICLE INFO

## ABSTRACT

Electricity markets are highly complex, involving lots of interactions and complex dependencies that make it hard to understand the inner workings of the market and what is driving prices. Econometric methods have been developed for this, white-box models, however, they are not as powerful as deep neural network models (DNN). In this paper, we use a DNN to forecast the price and then use XAI methods to understand the factors driving the price dynamics in the market. The objective is to increase our understanding of how different electricity markets work. To do that, we apply explainable methods such as SHAP and Gradient, combined with visual techniques like heatmaps (saliency maps) to analyse the behaviour and contributions of various features across five electricity markets. We introduce the novel concepts of SSHAP values and SSHAP lines to enhance the complex representation of high-dimensional tabular models.

## 1. Introduction

The main aim of this research article is to use Explainable AI to explore and understand wholesale electricity markets across different market economies.

The deregulation of electricity markets began in the 1980s, transforming traditional vertically integrated utilities into competitive structures to improve efficiency and foster innovation [1]. This process unbundled generation, transmission, and distribution, enabling independent entities to compete in electricity generation and retail. A key component of deregulated markets is the introduction of spot markets and forward markets, which provide platforms for trading electricity at different time horizons [2].

The spot market is primarily represented by the day-ahead market, where electricity is traded for delivery in hourly intervals during the following day, rather than the intraday market, which is more volatile [3]. This market plays a critical role in scheduling generation and consumption by matching supply and demand through a competitive bidding process.

Most markets operate under a merit order system, where the lowest-cost resources are dispatched first, progressively moving to higher-cost generators until demand is met, as in the European market [4]. Generators submit offers based on their marginal production costs, while buyers submit bids reflecting their willingness to pay for electricity. This ensures that electricity is supplied at the least cost while maintaining reliability. The market clearing price is determined by the marginal cost of the most expensive generator needed to meet the last unit of demand, a process known as a uniform price auction. In this system, all cleared generators receive the same price, regardless of their individual offers; it encourages suppliers to bid their true marginal costs, promoting efficiency in resource allocation [5].

In this paper, we analyse five distinct day-ahead markets; Germany, France, Belgium, Nord Pool and ComEd, that operate using the merit order mechanism with a uniform price auction.

### 1.1. Motivation and contributions

The size of the electricity market is immense and becoming even bigger. The electrification of the world's energy systems is one of the most important steps to limit climate change and reach net zero. The world electricity generation was 29.7 PWh in 2023 [6] and is projected to reach 95 PWh by 2050 in a scenario where all-purpose energy demand is matched with wind–water–solar electricity and heat supply [7].

As with many markets, price forecasting is a great tool for efficiently matching supply and demand. Furthermore, as electricity is an expensive commodity to store and storage is very limited, matching supply and demand is even more important for this market. The increasing complexity of electricity markets, driven by the integration of renewable energy and demand management, requires the development of robust AI models. However, the opacity of these models often limits their practical application. By focusing on explainability, we bridge this gap and provide actionable insights into the mechanisms driving electricity prices.

Our contributions are:

- We analyse five major markets with different XAI methods applied to DNN models.
- We introduce new tools called SSHAP values and SSHAP lines to explain and represent high-dimensional tabular DNN models.
- We bring new insights into the behaviours of these markets, like the impact of high load in France in the very early morning pushing down the prices in France and Belgium for the rest of the day.

### 1.2. Paper structure

The remainder of the paper is organised as follows. Section 2 performs a literature review of the current state of explainable methods in the electricity sector. Section 3 looks at the reference five datasets and models and Section 4 at the different explainable methods. Section 5 is the main section with the application of the methods to the five markets and the derived explanations. Finally, Section 6 provides a synthesis across the five markets and directions for further research in the field.

## 2. Literature review

Electricity price forecasting (EPF) has been a critical area of research due to its importance in the energy sector and the inherent complexity and volatility of deregulated electricity markets [8, 9]. Traditional econometric models, often referred to as white-box models, have been widely employed to predict electricity prices by leveraging fundamental market variables such as supply, demand, fuel prices, and weather conditions [10,11]. These models, while interpretable, are often limited in their capacity to capture nonlinearities and high-dimensional interactions present in electricity markets [12–14].

With the rise of machine learning, researchers have increasingly adopted its methods to address the challenges posed by traditional forecasting approaches in many sectors [15–18]. DNNs are capable of modelling complex dependencies and interactions in high-dimensional data, making them suitable for EPF. Studies have demonstrated their superior predictive accuracy compared to econometric methods and simpler machine learning models [19,20]. However, the black-box nature of DNNs presents significant challenges in interpretability, raising concerns about trust and transparency in critical decision-making processes.

To address the interpretability gap, eXplainable Artificial Intelligence (XAI) techniques have gained traction in recent years in the energy sector. XAI refers to the ability of AI models and systems to provide understandable and interpretable explanations for their decisions and predictions [21,22]. In the context of energy applications, XAI plays a crucial role in increasing transparency, trustworthiness, and accountability, particularly when AI is used in connection with high-stake activities [23]. Methods like SHAP (SHapley Additive exPlanations) and LIME (Local Interpretable Model-agnostic Explanations) have been widely used to provide insights into model predictions by quantifying the contribution of individual features to specific outcomes [24,25], for very diverse sectors and models like hotel booking cancellation prediction [26]. These tools enable stakeholders to understand the driving factors behind electricity price predictions, offering a balance between predictive power and interpretability.

We have found only very few research papers using XAI with EPF models. Two similar papers [27,28] use the density and local fit principles with the explanation methods Permutation Feature Importance [29,30] and SHAP to set a trust score to each forecast to determine how reliable it is. This in turn is used to improve the prediction accuracy of the EPF model. So the papers use SHAP mainly as intermediary variables in the stacking model rather than explanations for the electricity markets, which is our purpose. [31] uses a gradient-boosted tree model on the German market with twelve variables: five power system features and their ramps and fuel prices. The number of variables is small as the time dimension is only taken into account with the ramps rather than having 24 hourly variables for each power system feature. This allows the authors to use SHAP interaction values [32] between the different important variables and analyse their interactions. Instead, our models use more than one hundred variables.

The two papers closest to our research are [33,34] as they start from the same open-source platform built by [35] as we did. They both use the benchmark EPF models in [35] containing statistical and DNN types. The authors' contributions were to add more inputs to the datasets and most innovatively to apply the Kernel SHAP method to explain the models. [33] looks at Nord Pool and comes to very non-intuitive conclusions which are very different from ours; for example, the authors conclude that the prices three days earlier are the most important features of their DNN forecasting model. [34] looks mainly at the German market with multiple models and derives clear feature patterns and explanations like the importance of the Swiss electricity prices. Instead, our paper focuses on explanations only with deeper insights in multiple markets and the introduction of new explanation methods.

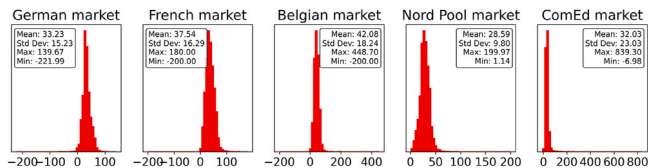

**Fig. 1.** Price histograms for the five markets.

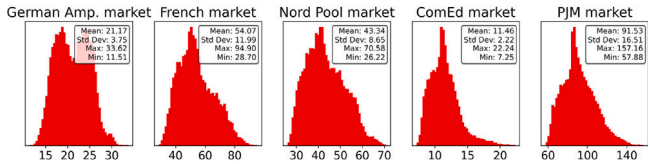

**Fig. 2.** Load histograms in GW.

## 3. Datasets and models

We implement our methods on top of the five open-access benchmark datasets and the reproducible open-source DNN models discussed in [35] and available in GitHub.[1]

### 3.1. Datasets

The datasets cover five day-ahead electricity markets, providing historical hourly electricity prices. Each dataset also includes two supplementary hourly time series: day-ahead forecasts for two significant exogenous factors, such as load demand and renewable generation. The represented markets are: Germany, France, and Belgium, operated by EPEX SPOT; Nord Pool, covering the Nordic countries; and Commonwealth Edison (ComEd), a zone located in northern Illinois, USA, operated by PJM. The data spans from September 5, 2012, to September 4, 2016, except for Nord Pool and ComEd, which begins on January 1, 2013. See the histograms of the prices and load volumes in Figs. 1 and 2.

The prices of each market have very distinct dynamics, i.e. they all have differences in terms of the frequency and existence of negative prices and price spikes. Germany, France and Belgium have very negative prices, down to −€200/MWh. It happened only once in France and Belgium while it is more frequent in Germany due to its high proportion of volatile renewable sources in its electricity mix [36]. Nord Pool has no negative prices and the least volatile prices with a standard deviation of €9.80/MWh thanks to its high proportion of hydropower [36]. ComEd has the most volatile prices with a standard deviation of $23.03/MWh, its minimum price is −$7/MWh but its maximum price is as high as $840/MWh.

### 3.2. Models

The models are day-ahead electricity price forecasts, with up to 241 inputs and 24 outputs covering each hour of the day-ahead. The DNN models are straightforward extensions of the traditional multilayer perceptron with two to four hidden layers with the Adam optimiser and the MAE (Mean Absolute Error) loss function [35]. In this paper, we use the DNN model with two hidden layers and 4 years of training data. The hyper-parameters are optimised using the tree-structured Parzen estimator [37], a sequential model-based optimisation algorithm within the family of Bayesian optimisation methods. The hyperparameters for the five DNN models are in Table 1.

---
[1] https://github.com/jeslago/epftoolbox.

In [35], the test sets are two years and for each date of the test set, a model is built using the four previous years as a training set. As we want to explain the models, rather than try to optimise them further, we only have one model per market and study them on their training sets.

The performance of the DNN models is shown in Table 2 (see performance metric definitions in [35]). As expected, they are better than in [35] as the metrics are applied to the training sets rather than the test sets.

## 4. Explainable methods

The most common methods for models using tabular data as input are Feature Importance methods containing a value for each feature for each instance. Each value indicates the importance of the feature for the prediction [21].

The DNN models are trained on normalised data for inputs and outputs to enhance their performance (see Table 1 for the specific normalisation for each model). However, when applying the explainable methods, we use denormalised outputs to quantify the absolute importance of the features rather than relative values.

Because these DNN models use smooth tabular data as input, all the standard Feature Importance methods gave very similar results, hence we use almost exclusively the SHAP method in the paper. We also use the Gradient method as it shows the directions of sensitivities in a more direct way than SHAP values. The Gradient method is simply to estimate the set of partial derivatives of the DNN model for each normalised input variable and each denormalised output variable at each instance.

### 4.1. SHAP method

SHAP covers Feature Importance methods based on the concept of the Shapley values from cooperative game theory [24]. The Shapley value is the average marginal contribution of a player across all possible player coalitions and it satisfies many desirable properties like efficiency where the sum of the Shapley values is equal to the overall game gain [38]. As we use the original values rather than the normalised values for the output when applying the SHAP method, the SHAP values have the same unit as the electricity prices, i.e. Euros or Dollars per MWh, which is useful. By assigning SHAP values to each feature, we quantify their contribution to the model's predictions across different instances.

The purpose of this paper is to explain the models rather than to explain the data, hence we use conditional expectation rather than marginal expectation when determining the Shapley values [39,40]. To estimate the conditional Shapley values, we use the Monte-Carlo sampling method developed by [41] and implemented in the GitHub SHAP library.

### 4.2. Heatmap representation

A heatmap is a visualisation technique for high-dimensional tabular datasets, where values are represented by colours, allowing the entire dataset to be captured in a coloured map. This visual tool is instrumental in identifying temporal patterns and relationships that may not be immediately apparent in numerical data.

For example, the French market has 120 inputs made of five super-variables: Price D-1, Price D-3, Load Forecast D, Generation Forecast D, and Generation Forecast D-1, each covering 24 inputs representing the 24 h of the day (120 = 5 × 24). The number of outputs is 24; so for each instance, the number of Feature Importance values (e.g. SHAP, Gradient) is 2880 (120 × 24).

So the SHAP heatmap for the French market is made of five square tables, each table being of dimension 24 × 24 (see Fig. 9(a)). The horizontal axis of each table represents the input hours and the vertical axis represents the output hours. So for example the point (i,j) in the Generation Forecast D-1 table indicates the impact of the generation forecast for the $i$th hour of day D-1 on the electricity price for the $j$th hour of day D (the day to be forecast).

**Table 1**
Hyperparameters of the five DNN models.

|  | Germany | France | Belgium | Nord Pool | ComEd |
| --- | --- | --- | --- | --- | --- |
| Number of neurons — Input layer | 217 | 120 | 121 | 144 | 120 |
| Number of neurons — First hidden layer | 329 | 233 | 205 | 274 | 299 |
| Number of neurons — Second hidden layer | 379 | 206 | 308 | 308 | 376 |
| Activation function | Softplus | Softplus | Softplus | Softplus | Selu |
| Dropout rate | 0.455 | 0.193 | 0.253 | 0.154 | 0.0079 |
| L1-regularisation factor | 0 | 0 | 0 | 0 | 0.0000306 |
| Network initialisation type | Glorot uniform | Glorot uniform | He normal | Lecun uniform | Lecun uniform |
| Input scaling type | Std | Arcsinh | Arcsinh | Median | Arcsinh |
| Output scaling type | Median | Std | Arcsinh | Std | Arcsinh |

**Table 2**
Performance measures for the five DNN models.

|  | Germany | France | Belgium | Nord Pool | ComEd |
| --- | --- | --- | --- | --- | --- |
| MAE | 2.47 | 3.32 | 5.29 | 1.23 | 3.01 |
| rMAE | 29.3% | 40.6% | 55.1% | 35.9% | 39.3% |
| sMAPE | 10.2% | 10.5% | 13.5% | 4.9% | 8.3% |
| RMSE | 3.83 | 5.31 | 11.59 | 2.37 | 9.76 |

Heatmaps can be used to represent the Feature Importance values for one instance or the average over the training set as in the next sections.

SHAP takes positive and negative values for each variable. For example for a linear model with independent variables, the SHAP value for the variable $x_i$ is simply the linear factor $a_i$ for the variable $x_i$ times $x_i$ minus the mean of $x_i$ [24]. So assuming that $a_i$ is positive, the SHAP value of $x_i$ is positive when $x_i$ is above its mean and negative otherwise. So heatmaps of average SHAP values have to be done with absolute value otherwise they would cancel each other out when averaging.

*4.3. Innovative representation with dimension reduction: SSHAP values and SSHAP lines*

A key feature of SHAP values is the efficiency property [24]: for any model m, the value of the model on an instance $x$ minus the mean value of m is equal to the sum of the SHAP values on $x$ for its n variables:

$$m(x) - E(m(X)) = \sum_{i=1}^{n} SHAP_i(x) \qquad (1)$$

For any partition of the set of variables into subsets, we call these subsets super-variables and for each subset or super-variable j, we define its Feature Importance value called SSHAP value (for Super SHAP value) as the sum of the SHAP values of the subset variables.

Thanks to the partition's properties of exclusivity and completeness, the sum through the subsets of their SSHAP values is equal to the total sum of the SHAP values. So we still have the efficiency property for the super-variables:

$$m(x) - E(m(X)) = \sum_{i=1}^{p} SSHAP_i(x) \qquad (2)$$

This is very useful for high-dimensional models to be able to represent them with less dimensions and it is different to applying SHAP to a set of variables like superpixels where only one SHAP value is calculated per superpixel.

For these EPF models, we define the super-variables simply by aggregating the variables through the 24-h input spectrum. For example, the Nord Pool model has 144 variables that give six super-variables covering 24 h: Price D-1, Price D-2, Load Forecast D, Load Forecast D-1, Wind Generation Forecast D and Wind Generation Forecast D-1. We also use the larger super-variables Price, Load Forecast and Wind Generation Forecast where Price aggregates Price D-1 and Price D-2, Load Forecast aggregates Load Forecast D and Load Forecast D-1 and Wind Generation Forecast aggregates Wind Generation Forecast D and in Wind Generation Forecast D-1 in the case of the Nord Pool market. With the three core super-variables, for example Load Forecast, all the offsetting between Load Forecast D and D-1 has gone, and SSHAP values represent the overall net impact of Load Forecast. Also if the heatmap reveals a significant shift for one super-variable, the super-variable can be split into two super-variables. For example, Load Forecast D was split into Load Forecast D H0-H4 and Load Forecast D H5-H23 in the French market because the gradients are almost all negative before 5 am and almost all positive after 5 am (see Fig. 9(b)).

We define SSHAP line for each super-variable as the weighted average of SSHAP values using a Gaussian kernel (with bandwidth of 5) against the actual price in the x-axis. This new method highlights not only the signs of SSHAP values but also their dependence on the price, see for example Fig. 11 for the French market. If the model is good and the average forecast price is close to the average actual price for all price bands, then, we can derive from Eq. (2) that the sum of the SSHAP lines should be almost a straight line with a slope of 1:

$$x - E(m(X)) = y \qquad (3)$$

**5. Explanations for five markets**

In this section, we present the results of applying the methodology outlined in Section 4 to the black-box DNN model used to forecast each market, one model calibrated for each market. [35] has shown that this model outperforms other approaches, providing the most accurate predictions of price. Our goal is that by applying these methods to this model, we can understand what the model has learned about how the market works and increase our own understanding of the market. As a counterfactual it may be helpful to consider a naïve forecast, which uses yesterday's price to predict today's price at the same time, i.e. today's price at 2 pm equals yesterday's price at 2 pm. When we use the explainable methods to explore what the black box has learned, we see far more complex dynamics. The naïve model is used in many papers as an efficient benchmark through the rMAE metric, i.e. the ratio of the MAE of the studied model over the MAE of the naïve model [11,42].

*5.1. The German market*

The dataset comprises the day-ahead forecast price, the day-ahead zonal load forecast in the TSO Amprion zone and the aggregated day-ahead wind and solar generation forecasts in the zones of the 3 largest TSOs (Amprion, TenneT, and 50 Hz). The German market is a large market in Europe which has made major investments in renewable generation, which makes up a significant portion of its generation mix [36]. Confirming that these exogenous variables are a good fit for the German market, [31] showed that load, solar and wind generations were the most important features for this market.

We look at how the model has used the features provided to make predictions, considering price, load and renewable forecasts separately for clarity.

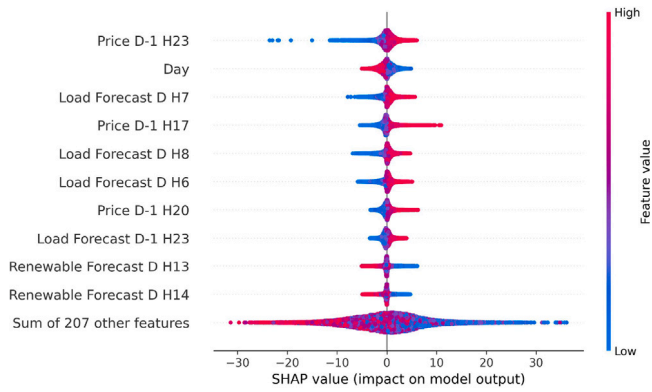

**Fig. 3.** SHAP beeswarm plot for the German market.

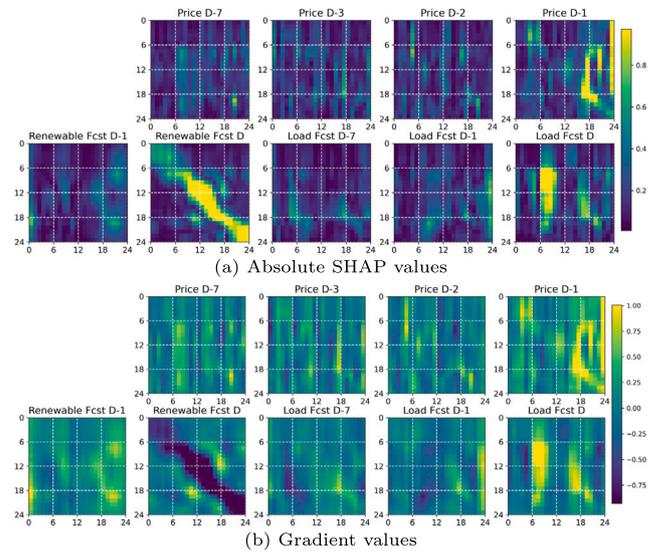

**Fig. 4.** Heatmaps for the German market.

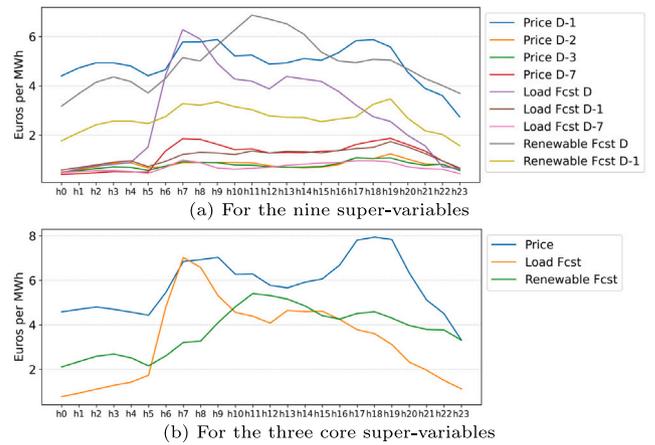

**Fig. 5.** Mean absolute SSHAP values throughout the day for the German market.

*5.1.1. Price variables*

From the beeswarm plot[2] in Fig. 3 we see that Price D-1 H23 is the most important feature overall with an average absolute value of €1.01/MWh, which is relatively low when compared to the price's standard deviation of €15.23/MWh. This means that the model uses a large number of variables to forecast prices. Price D-1 H17 and H20 also rank highly and are influential in the model. There is a general behaviour that high prices in these hours are associated with a higher average price forecast, however, we see that the effect varies in strength for H23 and H17, indeed high H17 values are the most informative feature for higher prices, while lower prices are more influenced by low D-1 H23. Similarly, H20 high values lead to higher predictions. We can untangle this behaviour more clearly using the SHAP heatmap in Fig. 4(a) for Price D-1. The 'column effect' we see in the images illustrates that Price D-1 H23 is influential in predicting across hours H0–H17. This is an interesting pattern compared to a naïve baseline of matching yesterday's hour to today's hour. Price D-1 H17 also exhibits this 'column' behaviour, being a prominent feature for predicting the prices in H9–H18. We see the model behave more similarly to this naïve 'hour matching' approach in the hours H18–H23, i.e. the model has learned that the evening peak prices from yesterday are more useful for forecasting today's evening peak prices, particularly H18 and H19.

However, we do see a 'horizontal' effect where there is equal weight given to Price D-1 H16–H19 to predict H18. This suggests an 'averaging' effect showing how the model is aggregating information from nearby hours to improve accuracy. We do also see a pattern of the morning peak being identified with Price D-1 H3–H5 important features for forecasting price H0–H5. The model seems to have learned distinct behaviours for different time periods. A particularly interesting hour is Price D H11, where we see the model uses Price D-1 H17, H20 and H23 almost equally to make this forecast. This shows a complex interaction between different time periods, which is very different from a naïve baseline hour matching.

The features Price D-2, Price, D-3 and Price D-7 are significantly less influential than D-1 however we do still see patterns of usage and again this column effect which highlights the importance of specific hours, for example, Price D-2 H4 affecting predictions for H0–H10. This suggests that the model has found ways of using these features, in combination with Price D-1.

Fig. 5 show the absolute importance of the 9 super-variables and the 3 core super-variables to the price forecasts throughout the day, averaging the hourly absolute SSHAP values across the training set. We see that Price D-1 is consistently ranked highly by the model throughout the day but the most important feature in two distinct time periods, H0–H6 and H16–H19. These periods line up well with the evening peak and the morning trough before the morning peak. Interestingly of the other price variables we see Price D-7, from a week ago, is ranked more important than Price D-2 and D-3 from H6 onwards throughout the rest of the day suggesting the model has learned a weekly effect for these hours. Although still significantly less influential than Price D-1.

In Fig. 6 we see the importance of features when predicting prices depending on the forecast price actual values. At the plot's highest price of €75/MWh, we see that Price D-1 is the most influential feature with an average SSHAP value of +€11/MWh, compared to +€2.5/MWh for Price D-7 and nothing for the other price variables. This suggests that high prices are predicted by Price D-1 and D-7. However, for low prices, we see that Price D-1 is second to Renewable Forecast D and that other price features are of little usage.

*5.1.2. Load Forecast*

Fig. 3 shows that the model uses Load Forecast D H7, H8 and H6 and interestingly Load Forecast D-1 H23 as influential features, where low values are indicative of low prices and high values high prices. It is surprising that D-1 H23 is selected over more recent features from Load Forecast D. When we consider Fig. 4(a) we see that Load Forecast D-1 H23 has this 'column' behaviour and is influential as a feature in predicting the price across hours H7–H22. Load Forecast D H6–H8 have a more concentrated effect on prices in H6–H16. This is further emphasised in Fig. 5(a) where we see that Load Forecast D is the most influential super-variable in the hours H7–H8 and maintains a high relative importance up to H17. The complex temporal behaviour exhibited by the price variable is also visible in the load forecast with both the column effect and the horizontal averaging effect being exhibited, for example, the price forecast at H7 is influenced by Load Forecast D H6, H7 and H8 equally and then to a lesser degree by H9, H10, H4 and others in Fig. 4(a).

---

[2] https://shap.readthedocs.io/en/latest/example_notebooks/api_examples/plots/beeswarm.html.

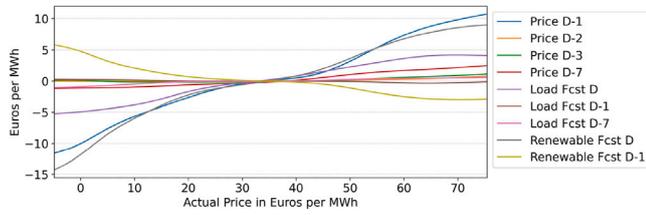

**Fig. 6.** SSHAP lines for the German market.

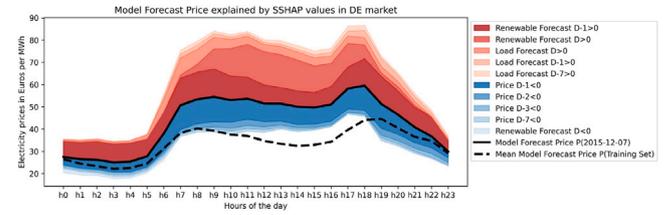

**Fig. 7.** SSHAP explanation for the forecast on 7 December 2015.

From Fig. 6 we see that Load Forecast D has negative SSHAP values for low and negative prices and positive values for high prices, as expected, but in both cases less influential than renewable forecast and price.

### 5.1.3. Renewable Forecast

In Fig. 3 we see that the two renewable forecast features ranked have opposite signs to the other variables indicating that high renewable generation forecasts for these hours are associated with lower prices and lower values with higher prices. The importance of the renewable forecast is highlighted in Fig. 5(a) ranking the averaged effect of the Renewable Forecast D as the most influential feature in the period H10–H15 and in Fig. 6 as the most important feature when predicting negative prices and second most important factor when predicting high prices. We also see that Renewable Forecast D-1 is ranked third in the periods H0–H5 and H18–H23, being more influential than Load Forecast D in these periods. Fig. 4 help to untangle the temporal complexity of the usage of this feature, and we see a striking pattern that is very different from both price and load forecast heatmaps. Renewable Forecast D exhibits a strong diagonal profile with negative gradients with particular emphasis on the hours from 8 am onwards, reinforcing the trend in Fig. 5(a). However, the model has learned to match the hour of the renewable forecast to the forecast price's while also using nearby hours. This may be a way of averaging the uncertainty in the renewable forecast to get more accurate predictions.

The influence of Renewable Forecast D-1 in Fig. 6 is in the opposite direction to Renewable Forecast D, a somewhat surprising result that suggests a complex offsetting interaction, between. The SSHAP line for the core super-variable Renewable Forecast is simply the sum of the two SSHAP lines for Renewable Forecast D and D-1, so that the importance of Renewable Forecast is significantly less than the Price's for low and high prices. This can also be seen in Fig. 5(b) where the impact of Renewable Forecast is around €2/MWh lower than Renewable Forecast D in Fig. 5(a) because of the offsetting with Renewable Forecast D-1. This is an offsetting on average when the renewable forecasts have about the same value for the day and the previous day. Instead, if the renewable forecast is high for the previous day and low for the day, SSHAP values for both Renewable Forecast D-1 and Renewable Forecast D push the prices up as in the case study on 7 December 2015 in Fig. 7, peaking at €60/MWh at H18.

### 5.2. The French market

The dataset comprises the day-ahead forecast price, the day-ahead load forecast and generation forecast. The French model utilises 5 super-variables, Price D-1, D-3, Load Forecast D, and Generation Forecast D and D-1, with 24 hourly variables used for each.

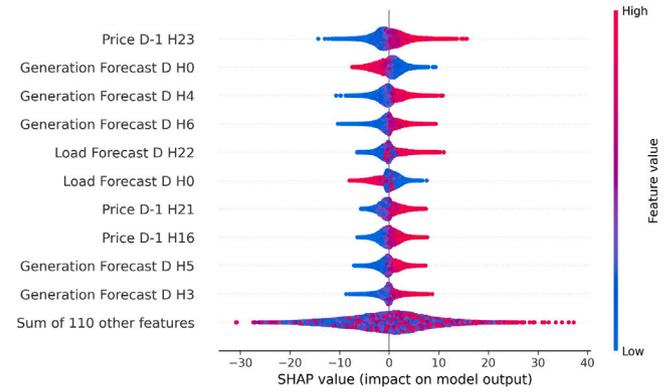

**Fig. 8.** SHAP beeswarm plot for the French market.

### 5.2.1. Price variables

Fig. 8 highlights Price D-1 H23, H21 and H16 as influential features where a high value predicts a high price and a low value is predictive of a lower price. H23 is the most influential of all in either low or high direction, with an average absolute value of €2.23/MWh, which is not insignificant when compared to the price's standard deviation of €16.29/MWh. Fig. 10(b) supports this effect where Price is ranked highest by far as the most predictive feature throughout the 24-h day. Interestingly Price D-3 is ranked lowest in Fig. 10(a) and is flat throughout the day, and Price D-1 is almost the same as Price. For extreme prices, we see that at both ends, high and low, Price D-1 is a very important feature in Fig. 11, with D-3 again having very little effect. For example, Price D-1 explains +€15/MWh on average from the mean price €38 MWh (see Eq. (2)) when Price D is around €80/MWh and almost −€10/MWh when Price D is nil. Fig. 9(a) helps us untangle the use of price features across different hours and we see interesting behaviour, particularly in D-1. The usage of H23, H22 and H21 exhibits a pronounced column effect where these 3 variables are combined to predict all hours of the day. We also see this column effect for H16. H16 in price D-3 is also prominent suggesting that this is a peak hour of interest. The price variables exhibit a recency focus on H21–H23 and a peak H16. This again contrasts with a naïve curve matching approach which would present as a strong diagonal in Fig. 9 for Price D-1 and D-3.

### 5.2.2. Load Forecast

Fig. 8 contains Load Forecast D H22 and H0 as important variables, interestingly they have opposite signs associated, indicating a complex relationship. In fact, the Gradient heatmap in Fig. 9(b) reveals a clear cut for Load Forecast D with all gradients being negative before 5 am (for predictions after 5 am) and mostly positive after 5 am for the input. For this reason, we split the super-variable Load Forecast D into two super-variables Load Forecast D H0-H4 and Load Forecast D H5-H23.

These negative gradients mean that the higher the load forecast from midnight to 5 am, the lower the prices after 5 am, and vice versa. The potential reason is that early high demand forces some non-flexible electricity generators to be ramped up which cannot be ramped down easily soon after. Hence they can create an oversupply of electricity later in the day and push the prices down. This is particularly the case in the French market where most of the generators are nuclear power plants with low flexibility [36].

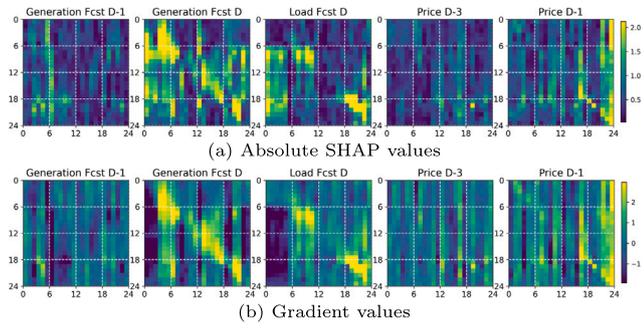

(a) Absolute SHAP values

(b) Gradient values

**Fig. 9.** Heatmaps for the French market. (For interpretation of the references to colour in this figure legend, the reader is referred to the web version of this article.)

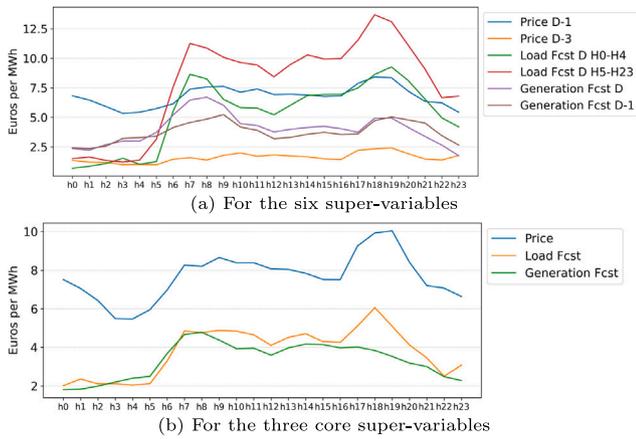

(a) For the six super-variables

(b) For the three core super-variables

**Fig. 10.** Mean absolute SSHAP values throughout the day for the French market.

Fig. 11 suggests that load forecast is predictive of the extremes in price particularly high prices. Also it shows that Load Forecast D H0-H4 offsets Load Forecast D H5-H23 by around half on average for high prices. For example, when Price D is at €80/MWh, Load Forecast D H5-H23 has explained +€20/MWh on average compared to −€11/MWh for Load Forecast D H0-H4. The offsetting happens when Load Forecast D is homogeneous throughout the day. Instead, if for example, Load Forecast D was lower than normal in the first five hours of the day and then higher than normal in the later hours, then the impact on the price would be very large as both super-variables would push the price up.

Considering Fig. 10 we see that the load forecast variable's importance is very dynamic across the day being very low for the hours H0–H5 and particularly important in the evening with a peak at H18. Fig. 9(a) highlights this varying usage of load forecast with Load Forecast D H11–H17 found not to be significant predictors, while the early morning values exhibit this column behaviour where they are useful in predicting price across a range of hours, although surprisingly not in their equivalent hours, i.e. load forecast at 3 am is not useful for predicting price at 3 am. However, it seems generation is more important for this early morning period.

#### 5.2.3. Generation Forecast

The generation variables selected in Fig. 8 are for Hours 0,3,4,5,6, suggesting that early morning forecast generation plays a strong role in setting price for the day. This may capture some aspect of the French generation mix, with nuclear not particularly flexible. With 5 out of the top 10 variables from Generation Forecast D we might expect it to be a very significant feature but Fig. 10(a) shows only a medium importance throughout the day with the most prominence in predicting prices in hours 6 am to 10 am, suggesting the morning peak is significantly affected by this variable. Generation Forecast D-1 is of similar but less importance throughout the day.

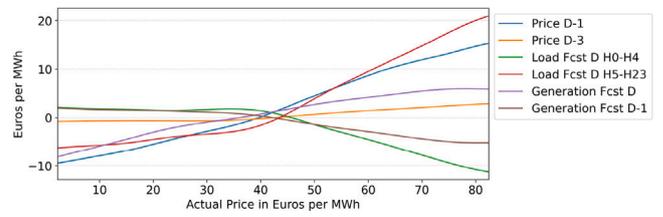

**Fig. 11.** SSHAP lines for the French market.

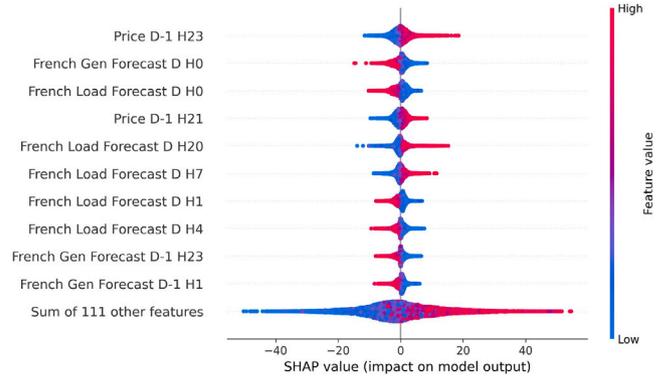

**Fig. 12.** SHAP beeswarm plot for the Belgian market.

4 out of the 5 variables show a positive impact on higher prices for high generation forecast values in Fig. 8, which is counterintuitive. Similarly, Generation Forecast D has a majority of positive gradients in Fig. 9(b).

The behaviour exhibited in Fig. 9(a) for generation is also complex with many hours contributing to the model, however we see column like behaviour again with the generation forecast in early hours significantly influencing prices throughout the day. We also see horizontal averaging behaviour where the generation forecast from multiple hours is being used to predict a specific price, for example, the forecast price for H7 and H8 is influenced by 8 h of generation forecast values. This highlights the model is doing something more sophisticated than curve matching from yesterday's generation forecast.

Fig. 11 shows that Generation Forecast D and D-1 are offsetting each other exactly on average for high prices. Instead, Generation Forecast D is important for predicting very low prices and it is hardly offset by Generation Forecast D-1.

### 5.3. The Belgian market

The Belgian market is tightly connected to the much larger French market and as such the Belgian model includes variables that are the French Load Forecast D and D-7, French Generation Forecast D and D-1 as well as Price D-1.

#### 5.3.1. Price variables

Fig. 12 highlights two price variables, for Price D-1 H23 and H21 as significant variables. We see that high values here predict high prices and vice versa for low values. Fig. 14(b) ranks the average effect of Price as the most important core super-variable across the day. Similarly, the feature is the most important at the extremes of both negative and positive prices in Fig. 15. Fig. 13(a) helps untangle more of the temporal variability in how yesterday's prices influence the model. We see a column effect on hours H23 and H21 which are used to influence predictions for almost all of the 24 h of the day. We also see contributions from more of the diagonal elements in the early morning and late evening. This highlights a 'recency bias' where the most recent information is more informative than the price from the corresponding hour, i.e. price D-1 H23 is a stronger predictor of Price D H9 than Price D-1 H9.

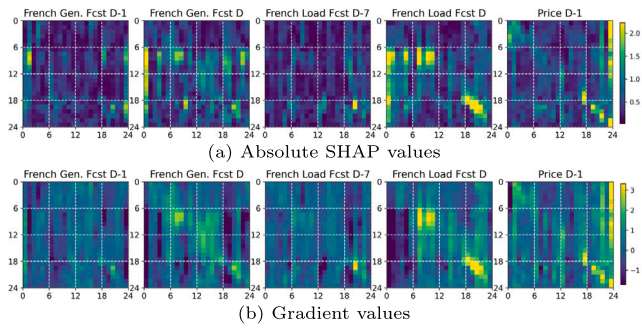

(a) Absolute SHAP values

(b) Gradient values

**Fig. 13.** Heatmaps for the Belgian market.

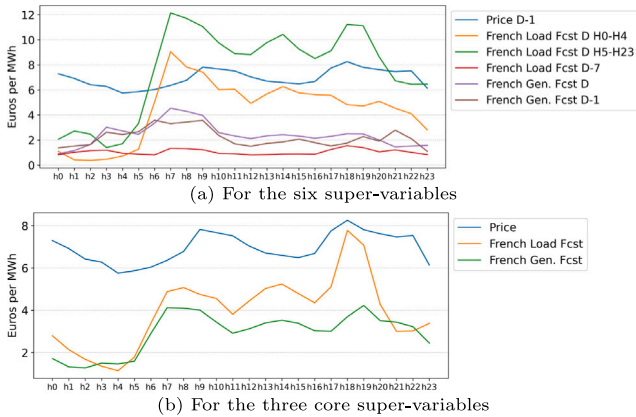

(a) For the six super-variables

(b) For the three core super-variables

**Fig. 14.** Mean absolute SSHAP values throughout the day for the Belgian market.

*5.3.2. French Load Forecast*

Five out of the top 10 variables in Fig. 12 are from the French Load Forecast D, none from the French Load Forecast D-7. They have opposite directions in sensitivity between the very early hours H0, H1 and H4 and the later hours H20 and H7. Similar to the French market, the Gradient heatmap in Fig. 13(b) reveals a clear cut with almost all gradients being negative before 5 am (for predictions after 6 am) and positive or nil after 5 am for the input. For this reason we split the super-variable French Load Forecast D into two super-variables French Load Forecast D H0-H4 and French Load Forecast D H5-H23.

Fig. 15 shows that French Load Forecast D H0-H4 offsets French Load Forecast D H5-H23 by around half on average for high prices. For example, when Price D is at €80/MWh, French Load Forecast D H5-H23 has explained +€12/MWh on average compared to −€5/MWh for French Load Forecast D H0-H4.

French Load Forecast D H5-H23 is the most important feature for most of the day in Fig. 14(a). French Load Forecast D H0-H4 is third and French Load Forecast D-7 is the least informative for most of the day instead. Fig. 13(a) helps unpack the temporal relationships of the load forecast feature and we see a columning effect with the hours H0-H4 particularly useful predictors of price throughout the day, and instead with a section of non-contribution in the region H11-H16. It appears these hours were found not to be as useful by the model.

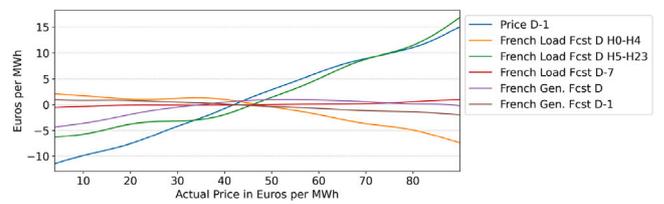

**Fig. 15.** SSHAP lines for the Belgian market.

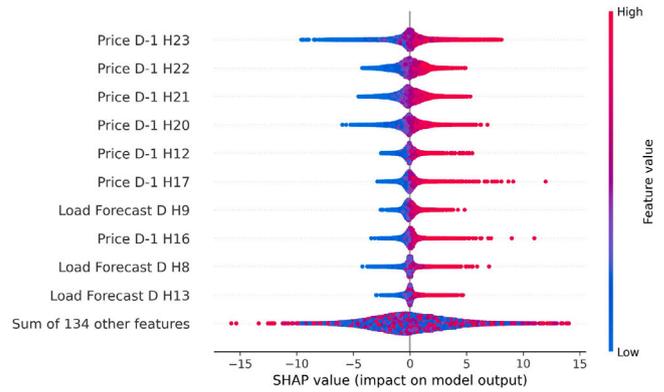

**Fig. 16.** SHAP beeswarm plot for the Nord Pool market.

*5.3.3. French Generation Forecast*

Fig. 12 shows the model make use of features from French Generation Forecast D and D-1, with H0 from D and H23 and H1 from D-1 selected as important variables. The signs are consistent with high generation leading to low prices and vice versa. Fig. 14(a) ranks both super-variables as similarly mildly important although Generation Forecast D is more influential as expected with both showing an increase in importance in the morning peak hours from H5 to H10. They are poor predictors of high prices in Fig. 15, but Generation Forecast D is a useful predictor of low prices which is interesting.

*5.4. The Nord Pool market*

The dataset comprises the day-ahead forecast Nordic system price, a theoretical, unconstrained market price calculated for the entire Nord Pool area. Besides the prices, the dataset comprises the day-ahead load forecast and wind power forecast for the entire Nord Pool area.

*5.4.1. Price variables*

Fig. 16 is striking in that 7 of the top 10 variables are price variables, with a strong cluster of H20-H23 and then H12, H16 and H17. The signs for all features are consistent with high values indicating higher predictions. No variables from Price D-2 feature and this is reinforced in Fig. 18(a) with the model's stark preference for D-1 over D-2. D-1 is the most significant predictor across the day. It is also the most influential predictor by far for both high and low extreme prices in Fig. 19. Fig. 17(a) shows a strong column effect for Price D-1 with H20-H23 used as predictors for all hours of the day.

*5.4.2. Load Forecast*

Three Load Forecast D variables feature in Fig. 16, H8, H9 and H13. These hours contrast with the model's preference for use of evening based price features. Indeed when we consider Fig. 18(a) we see a large peak in the importance of Load Forecast D and D-1 from H6 to H9 and then a peak in the evening again.

Fig. 18(b) reveals that when the SSHAP values are netted between Load Forecast D and D-1, the values are much lower by more than half. This means that Load Forecast D and D-1 are mainly offsetting each other. The correlation of their SHAP values is -16% only but the correlation of their SSHAP values is -92% instead! This means that the offsetting is not done simply between Load Forecast D-1 and Load Forecast D for the same hours but for different hours throughout the day. This shows that the model is learning complex temporal dynamics, which is very interesting. Fig. 19 shows that Forecast D and Load Forecast D-1 are almost symmetric, which is consistent with their high offsetting. For load forecast to have a significant impact on high prices, the load should be low on the previous day and high on the day, to give a ramp-up effect.

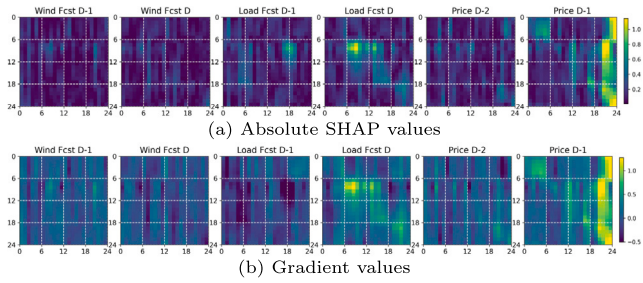

Fig. 17. Heatmaps for the Nord Pool market.

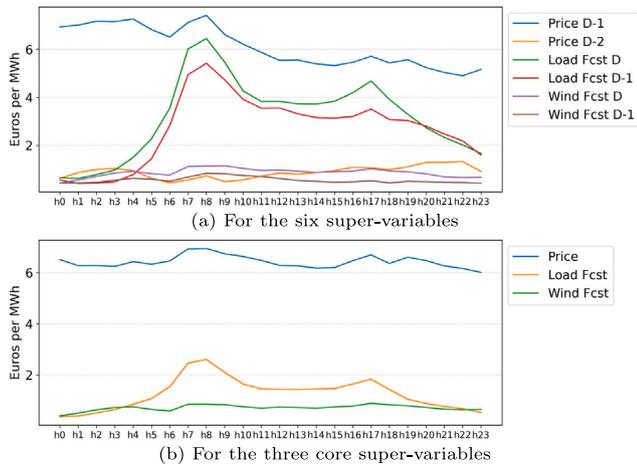

Fig. 18. Mean absolute SSHAP values throughout the day for the Nord Pool market.

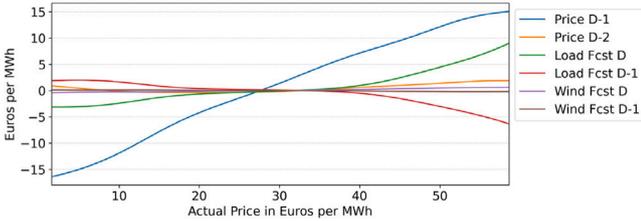

Fig. 19. SSHAP lines for the Nord Pool market.

#### 5.4.3. Wind Generation Forecast

All figures indicate that the wind generation forecasts are not found to be significant predictors by the model which is interesting. Nord Pool is dominated by significant hydropower generation which accounts for half of generation and nuclear accounts for a third [36] suggesting wind is not a significant factor in determining price at its current market share.

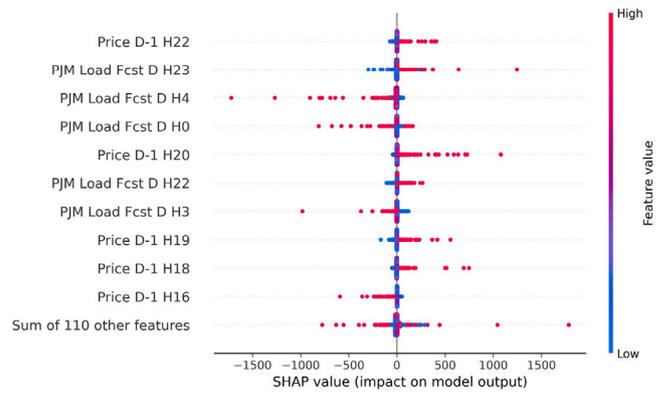

Fig. 20. SHAP beeswarm plot for the ComEd market.

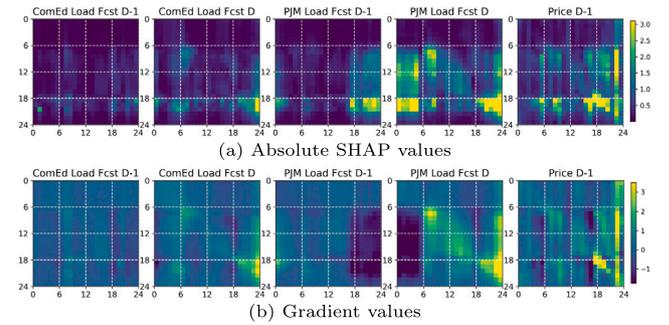

Fig. 21. Heatmaps for the ComEd market.

#### 5.5. The ComEd market

The dataset contains the day-ahead prices of the ComEd zone, the day-ahead ComEd load forecast and the day-ahead load forecast for all the 13 states covered by PJM. The ComEd model utilises the super-variables Price D-1, PJM Load Forecast D, D-1 and ComEd Load Forecast D and D-1.

##### 5.5.1. Price variables

Fig. 20 shows some extreme SHAP values for the top 10 variables of a few hundred dollars per MWh. This seems abnormal even if ComEd prices reached levels as high as $840/MWh.

Price D-1 H22 is the most important variable out of the 120 variables. This is the last price available as we use local time in the datasets, which is the Central Time (CT) for Illinois while PJM operates with the Eastern Time (ET). So the last day-ahead contract spans from 11 pm to midnight ET, which corresponds to 10 pm to 11 pm CT. Its average absolute SHAP value is $2.51/MWh, which shows that the percentage of extreme SHAP values is very small.

There are four other price variables in Fig. 20, Price D-1 H20, H19, H18 and H16. High prices lead to higher day-ahead forecast prices except for H16, which is surprising. H16 seems to be unique with negative gradients according to Fig. 21(b).

Figs. 21(a) and 22(a) show that the SHAP values are much higher in the afternoon than in the morning and the price forecasts depend on many price variables. In particular, price forecast H18–H20 rely on most input price variables to be predicted according to Fig. 22(a). This confirms that the model is complex in using many inputs for each instance.

Fig. 23 shows that Price D-1 is an important feature to predict high price. For example for Price D at $100/MWh, the average SSHAP value for Price D-1 is +$33/MWh. Price D-1 is less influential for low prices.

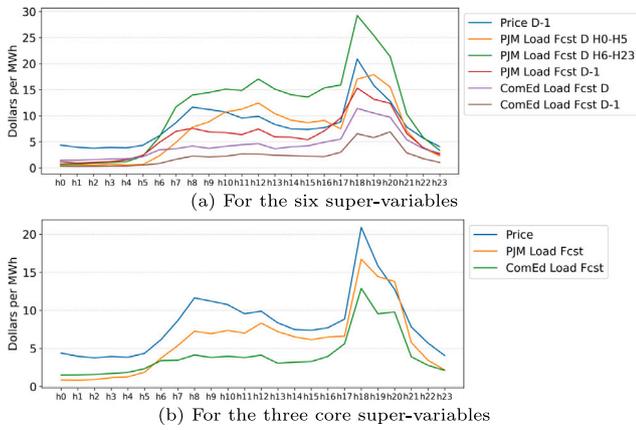

(a) For the six super-variables

(b) For the three core super-variables

**Fig. 22.** Mean absolute SSHAP values throughout the day for the ComEd market.

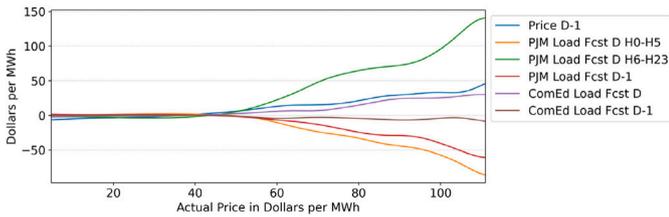

**Fig. 23.** SSHAP lines for the ComEd market.

*5.5.2. PJM Load Forecast*

PJM Load Forecast D also has five important variables in Fig. 20 with some extreme values. Two are from the very late hours H23 and H22 and their high values are associated with high price forecasts. Three are from the very early hours H4, H0 and H3 and their sensitivities are opposite to the late hours'. This is confirmed from the Gradient heatmap in Fig. 21(b) showing negative gradients for almost all Load Forecast D for H0–H5 relative to forecast prices for H6–H23. In fact PJM Load Forecast D-1 also has negative gradients for H17–H23 relative to forecast prices for H6–H23, so overall the gradients are negative from 5 pm the day before to 6 am on the day. For this reason, we split the super-variable PJM Load Forecast D into PJM Load Forecast D H0–H5 and PJM Load Forecast D H6–H23.

Figs. 22(a) and 23 show that PJM Load Forecast D H6-H23 is the most important variable, but it is significantly offset by both PJM Load Forecast D H0–H5 and PJM Load Forecast D-1. For example for Price D at $100/MWh, the average SSHAP value for PJM Load Forecast D H6-H23 is as much as +$96/MWh according to Fig. 23 but it is offset by PJM Load Forecast D H0–H5 and PJM Load Forecast D-1 which average SSHAP values are −$58/MWh and −$40/MWh respectively.

Fig. 21(a) shows that there are many PJM load forecast variables influencing the price forecast for the period H18-H20. This is the reason why the super-variables' importance peak during the period H18-H20 in Fig. 22.

*5.5.3. ComEd Load Forecast*

ComEd load forecast does not have any variable in the top 10 variables in Fig. 20 and its heatmaps in Fig. 21 do not show any intense areas except when forecasting for the hours H18-H20. But Fig. 23 shows that ComEd Load Forecast D is important for forecasting high prices, in particular because it is hardly offset by ComEd Load Forecast D-1, which has a very small impact on price forecast. For example for Price D at $100/MWh, the average SSHAP value for ComEd Load Forecast D is +$25/MWh according to Fig. 23 and it is offset by ComEd Load Forecast D-1 by only −$4/MWh.

## 6. Synthesis, discussion and future work

Here is the synthesis of the market behaviours noticeable in the five markets:

- The last price available is the most important variable to predict the day-ahead prices for the five markets. The models show a preference for more recent variables over a 'price curve matching' approach which is a more natural human approach and used quite often in state of the art. Instead of the model predicting the price at 10 am using yesterday's price at 10 am, it finds better predictions using yesterday's most recent prices. This is a very interesting finding that sheds light into the inner workings of the black-box models, and highlights a key difference with more intuitive approaches.
  Still, the impact of the last price available is always moderate, which shows that the models are complex in using many of their 100+ variables. The last price available is very influential when forecasting in the early hours of the day ahead and its importance fades with the later hours. For the late hours, the 'price curve matching' does become prominent, as it has the matching hour benefit and the prices are close in time to the last price available.
- The results show a strong difference between the German market where the price is highly influenced by renewable generation and France, Belgium, Nord Pool and ComEd, which are more 'load' driven markets. Germany is more advanced in its renewable development and it may be that this is an artefact of the generation mix, but the model has learned interesting dynamics about the drivers of price. Load's importance is very dynamic throughout the day. It is extremely low in the first hours and jumps to its maximum in the morning during its ramp-up. Then it decreases throughout the day.
- Neural networks' key strength is their ability to do feature transformation inside the black box and it is clear there is significant complex interactions between the variables being performed where during different parts of the day different features are more predictive.
- In three markets, France, Belgium and ComEd, the impact of high load forecast in the very early morning is to push the prices down for the rest of the day. The Gradient heatmaps have a very distinct rectangle area of negative values reflecting this, and this is translated in the SSHAP curves for Load D H0-H4 and H5-H23 going in opposite directions and offsetting each other on average.
- Some exogenous lagged super-variables, like Renewable Forecast D-1, primarily counterbalance the impact of their corresponding current super-variables, like Renewable Forecast D, when their values are similar between the two days. However, if these values change significantly from one day to the next, their effects accumulate instead, leading to a ramp-up effect.

The five markets exhibit distinct characteristics, as discussed in the paper and highlighted by three complexity measures set in Table 3. Nord Pool stands out as the simplest market, with 'Price D-1' being the predominant super-variable, and it registers the lowest complexity across all three measures. This could explain the high performance of the Nord Pool model (MAE is €1.23/MWh only). The German market also displays all sensitivities as expected, but it is more complex with the two exogenous variables playing a significantly larger role. In contrast, the French and Belgian markets, which are similar to each other, feature heterogeneous heatmaps that indicate numerous variables offsetting each other, suggesting intricate temporal dynamics. The ComEd market presents slightly more homogeneous heatmaps than the French and Belgian markets; however, the model exhibits very large values in gradients and SHAP values, reflecting a high degree of non-linearity.

This paper provides an overview of explainable methods applied to neural network models for electricity price forecasting across five

Table 3

Non-linearity measures the average standard deviation of the gradients (this would be zero for a linear model). Non-homogeneity measures the average of the absolute differences between neighbouring cells of the SHAP heatmap. Number of important variables per hour is the number of cells of the SHAP heatmap above the threshold €0.5/MWh divided by 24 (the number of hourly forecasts).

|  | Germany | France | Belgium | NordPool | ComEd |
|---|---|---|---|---|---|
| Non-linearity | 0.182 | 0.635 | 0.513 | 0.102 | 1.824 |
| Non-homogeneity | 0.089 | 0.325 | 0.278 | 0.075 | 0.237 |
| Nb of important variables per hour | 19.7 | 67.5 | 51.0 | 7.4 | 42.4 |

markets. By using SHAP values and gradients, along with SSHAP and heatmap visualisations, we obtained insights into the factors influencing electricity prices and their complex temporal dynamics. Our analysis revealed consistent important features in all five markets and also market specific features. The study highlights the value of XAI in understanding model behaviour, particularly in complex domains like electricity price forecasting.

The analysis provides actionable insights for market operators and policymakers. The ability to identify dominant features and their temporal patterns enhances transparency and fosters trust in AI-driven decision-making processes.

The level of complexity of each market can be derived from the volatility of its electricity price (Fig. 1) and the accuracy performance of its forecasting model (Table 2). For each market our XAI methods succeeded in explaining where the complexity was coming from and it is summarised in Table 3.

Future work will explore integrating additional exogenous variables like the different sources of the electricity generation mix and the interconnector flow amounts between the neighbouring bidding zones. These additional variables will increase further the dimensionality of the model and make the SSHAP values and SSHAP lines even more relevant to understand the models. Other potential directions for further research are to analyse explanations for extreme prices in particular where the models tend to perform less well and to use the XAI methods to determine potential dysfunction of the models (e.g. how can the ComEd model create such extreme SHAP values).

**CRediT authorship contribution statement**

**Antoine Pesenti:** Writing – review & editing, Writing – original draft, Methodology, Formal analysis, Data curation, Conceptualization. **Aidan O'Sullivan:** Writing – original draft, Supervision, Conceptualization.

**Declaration of generative AI and AI-assisted technologies in the writing process**

During the preparation of this work the author(s) used ChatGPT and Copilot in order to improve the readability of some sections of the paper. After using this tool/service, the author(s) reviewed and edited the content as needed and take(s) full responsibility for the content of the publication.

**Declaration of competing interest**

The authors declare the following financial interests/personal relationships which may be considered as potential competing interests: Antoine Pesenti reports financial support was provided by EDF Energy R&D UK Centre Limited. If there are other authors, they declare that they have no known competing financial interests or personal relationships that could have appeared to influence the work reported in this paper.

**Acknowledgements**

This work was supported by EDF Energy R&D UK Centre Limited and EPSRC under Grant EP/V519625/1.

**Data availability**

The data that support the findings of this study are available at https://zenodo.org/records/4624805 or can be accessed with the code from https://github.com/jeslago/epftoolbox.

**References**

[1] Joskow PL. Lessons learned from electricity market liberalization. Energy J 2008;29(2_suppl):9–42. http://dx.doi.org/10.5547/ISSN0195-6574-EJ-Vol29-NoSI2-3, URL https://journals.sagepub.com/doi/10.5547/ISSN0195-6574-EJ-Vol29-NoSI2-3.

[2] Wolak FA. Market design and price behavior in restructured electricity markets: An international comparison. In: Faruqui A, Eakin K, editors. Pricing in competitive electricity markets. Boston, MA: Springer US; 2000, p. 127–52. http://dx.doi.org/10.1007/978-1-4615-4529-3_8, URL http://link.springer.com/10.1007/978-1-4615-4529-3_8.

[3] Mayer K, Trück S. Electricity markets around the world. J Commod Mark 2018;9:77–100. http://dx.doi.org/10.1016/j.jcomm.2018.02.001, URL https://www.sciencedirect.com/science/article/pii/S2405851318300059.

[4] Gasparella A, Koolen D, Zucker A. The merit order and price-setting dynamics in European electricity markets. 2023, URL https://publications.jrc.ec.europa.eu/repository/handle/JRC134300.

[5] Cramton P, Stoft S. Why we need to stick with uniform-price auctions in electricity markets. Electr J 2007;20(1):26–37. http://dx.doi.org/10.1016/j.tej.2006.11.011, URL https://www.sciencedirect.com/science/article/pii/S1040619006001527.

[6] IEA. Electricity 2024: Analysis and forecast to 2026. 2024, URL https://iea.blob.core.windows.net/assets/18f3ed24-4b26-4c83-a3d2-8a1be51c8cc8/Electricity2024-Analysisandforecastto2026.pdf.

[7] Jacobson MZ. A solution to global warming, air pollution, and energy insecurity for 149 countries. 2023, URL https://web.stanford.edu/group/efmh/jacobson/Articles/I/149Country/24-WWS-149Countries.pdf.

[8] Maciejowska K, Uniejewski B, Weron R. Forecasting electricity prices. 2022, http://dx.doi.org/10.48550/arXiv.2204.11735, URL http://arxiv.org/abs/2204.11735. arXiv:2204.11735 [q-fin].

[9] Gil HA, Gómez-Quiles C, Gómez-Expósito A, Santos JR. Forecasting prices in electricity markets: Needs, tools and limitations. In: Sorokin A, Rebennack S, Pardalos PM, Iliadis NA, Pereira MVF, editors. Handbook of networks in power systems i. Berlin, Heidelberg: Springer; 2012, p. 123–50. http://dx.doi.org/10.1007/978-3-642-23193-3_5.

[10] Contreras J, Espinola R, Nogales F, Conejo A. ARIMA models to predict next-day electricity prices. IEEE Trans Power Syst 2003;18(3):1014–20. http://dx.doi.org/10.1109/TPWRS.2002.804943, URL https://ieeexplore.ieee.org/abstract/document/1216141. Conference Name: IEEE Transactions on Power Systems.

[11] Conejo AJ, Contreras J, Espínola R, Plazas MA. Forecasting electricity prices for a day-ahead pool-based electric energy market. Int J Forecast 2005;21(3):435–62. http://dx.doi.org/10.1016/j.ijforecast.2004.12.005, URL https://www.sciencedirect.com/science/article/pii/S0169207004001311.

[12] Aggarwal SK, Saini LM, Kumar A. Electricity price forecasting in deregulated markets: A review and evaluation. Int J Electr Power Energy Syst 2009;31(1):13–22. http://dx.doi.org/10.1016/j.ijepes.2008.09.003, URL https://linkinghub.elsevier.com/retrieve/pii/S0142061508000884.

[13] Weron R. Electricity price forecasting: A review of the state-of-the-art with a look into the future. Int J Forecast 2014;30(4):1030–81. http://dx.doi.org/10.1016/j.ijforecast.2014.08.008, URL https://linkinghub.elsevier.com/retrieve/pii/S0169207014001083.

[14] Laitsos V, Vontzos G, Paraschoudis P, Tsampasis E, Bargiotas D, Tsoukalas LH. The state of the art electricity load and price forecasting for the modern wholesale electricity market. Energies 2024;17(22):5797. http://dx.doi.org/10.3390/en17225797, URL https://www.mdpi.com/1996-1073/17/22/5797. Number: 22 Publisher: Multidisciplinary Digital Publishing Institute.

[15] Dong Y, Zhang L, Liu Z, Wang J. Integrated forecasting method for wind energy management: A case study in China. Processes 2020;8(1):35. http://dx.doi.org/10.3390/pr8010035, URL https://www.mdpi.com/2227-9717/8/1/35. Number: 1 Publisher: Multidisciplinary Digital Publishing Institute.

[16] Jiang P, Liu Z, Wang J, Zhang L. Decomposition-selection-ensemble prediction system for short-term wind speed forecasting. Electr Power Syst Res 2022;211:108186. http://dx.doi.org/10.1016/j.epsr.2022.108186, URL https://www.sciencedirect.com/science/article/pii/S0378779622003960.

[17] Jiang P, Liu Z, Zhang L, Wang J. Hybrid model for profit-driven churn prediction based on cost minimization and return maximization. Expert Syst Appl 2023;228:120354. http://dx.doi.org/10.1016/j.eswa.2023.120354, URL https://www.sciencedirect.com/science/article/pii/S0957417423008564.

[18] Dong Y, Sun Y, Liu Z, Du Z, Wang J. Predicting dissolved oxygen level using Young's double-slit experiment optimizer-based weighting model. J Environ Manag 2024;351:119807. http://dx.doi.org/10.1016/j.jenvman.2023.119807, URL https://linkinghub.elsevier.com/retrieve/pii/S0301479723025951.

[19] Khosravi A, Nahavandi S, Creighton D. Quantifying uncertainties of neural network-based electricity price forecasts. Appl Energy 2013;112:120–9. http://dx.doi.org/10.1016/j.apenergy.2013.05.075, URL https://www.sciencedirect.com/science/article/pii/S0306261913004881.

[20] Lago J, De Ridder F, De Schutter B. Forecasting spot electricity prices: Deep learning approaches and empirical comparison of traditional algorithms. Appl Energy 2018;221:386–405. http://dx.doi.org/10.1016/j.apenergy.2018.02.069, URL https://linkinghub.elsevier.com/retrieve/pii/S030626191830196X.

[21] Bodria F, Giannotti F, Guidotti R, Naretto F, Pedreschi D, Rinzivillo S. Benchmarking and survey of explanation methods for black box models. 2021, URL http://arxiv.org/abs/2102.13076. arXiv:2102.13076 [cs].

[22] Molnar C. Interpretable machine learning. Molnar; 2022, URL https://christophm.github.io/interpretable-ml-book/.

[23] Machlev R, Heistrene L, Perl M, Levy K, Belikov J, Mannor S, Levron Y. Explainable artificial intelligence (XAI) techniques for energy and power systems: Review, challenges and opportunities. Energy AI 2022;9:100169. http://dx.doi.org/10.1016/j.egyai.2022.100169, URL https://linkinghub.elsevier.com/retrieve/pii/S2666546822000246.

[24] Lundberg S, Lee S-I. A unified approach to interpreting model predictions. 2017, URL http://arxiv.org/abs/1705.07874. arXiv:1705.07874 [cs, stat].

[25] Ribeiro MT, Singh S, Guestrin C. "Why should i trust you?": Explaining the predictions of any classifier. In: Proceedings of the 22nd ACM SIGKDD international conference on knowledge discovery and data mining. San Francisco California USA: ACM; 2016, p. 1135–44. http://dx.doi.org/10.1145/2939672.2939778, URL https://dl.acm.org/doi/10.1145/2939672.2939778.

[26] Liu Z, De Bock KW, Zhang L. Explainable profit-driven hotel booking cancellation prediction based on heterogeneous stacking-based ensemble classification. European J Oper Res 2025;321(1):284–301. http://dx.doi.org/10.1016/j.ejor.2024.08.026, URL https://www.sciencedirect.com/science/article/pii/S0377221724006696.

[27] Heistrene L, Machlev R, Perl M, Belikov J, Baimel D, Levy K, Mannor S, Levron Y. Explainability-based trust algorithm for electricity price forecasting models. Energy AI 2023;14:100259. http://dx.doi.org/10.1016/j.egyai.2023.100259, URL https://linkinghub.elsevier.com/retrieve/pii/S2666546823000319.

[28] Heistrene L, Belikov J, Baimel D, Katzir L, Machlev R, Levy K, Mannor S, Levron Y. An improved and explainable electricity price forecasting model via SHAP-based error compensation approach. IEEE Trans Artif Intell 2024;1–11. http://dx.doi.org/10.1109/TAI.2024.3455313, URL https://ieeexplore.ieee.org/document/10669041. Conference Name: IEEE Transactions on Artificial Intelligence.

[29] Breiman L. Random forests. Mach Learn 2001;45(1):5–32. http://dx.doi.org/10.1023/A:1010933404324.

[30] Fisher A, Rudin C, Dominici F. All models are wrong, but many are useful: Learning a variable's importance by studying an entire class of prediction models simultaneously. J Mach Learn Res 2019;20(177):1–81, URL http://jmlr.org/papers/v20/18-760.html.

[31] Trebbien J, Rydin Gorjão L, Praktiknjo A, Schäfer B, Witthaut D. Understanding electricity prices beyond the merit order principle using explainable AI. Energy AI 2023;13:100250. http://dx.doi.org/10.1016/j.egyai.2023.100250, URL https://linkinghub.elsevier.com/retrieve/pii/S2666546823000228.

[32] Lundberg SM, Erion GG, Lee S-I. Consistent individualized feature attribution for tree ensembles. 2019, http://dx.doi.org/10.48550/arXiv.1802.03888, URL http://arxiv.org/abs/1802.03888. arXiv:1802.03888 [cs].

[33] Mascarenhas MM, Amelin M, Kazmi H. Bridging accuracy and explainability in electricity price forecasting. In: 2024 20th international conference on the European energy market. EEM, 2024, p. 1–6. http://dx.doi.org/10.1109/EEM60825.2024.10608857, URL https://ieeexplore.ieee.org/document/10608857/?arnumber=10608857. ISSN: 2165-4093.

[34] Tschora L, Pierre E, Plantevit M, Robardet C. Electricity price forecasting on the day-ahead market using machine learning. Appl Energy 2022;313:118752. http://dx.doi.org/10.1016/j.apenergy.2022.118752, URL https://linkinghub.elsevier.com/retrieve/pii/S0306261922002057.

[35] Lago J, Marcjasz G, De Schutter B, Weron R. Forecasting day-ahead electricity prices: A review of state-of-the-art algorithms, best practices and an open-access benchmark. Appl Energy 2021;293:116983. http://dx.doi.org/10.1016/j.apenergy.2021.116983, URL https://linkinghub.elsevier.com/retrieve/pii/S0306261921004529.

[36] Zakeri B, Staffell I, Dodds PE, Grubb M, Ekins P, Jääskeläinen J, Cross S, Helin K, Castagneto Gissey G. The role of natural gas in setting electricity prices in Europe. Energy Rep 2023;10:2778–92. http://dx.doi.org/10.1016/j.egyr.2023.09.069, URL https://www.sciencedirect.com/science/article/pii/S2352484723013057.

[37] Bergstra JS, Bardenet R, Bengio Y, Kégl B. Algorithms for hyper-parameter optimization. Adv Neural Inf Process Syst 2011;24. URL https://core.ac.uk/download/pdf/46766638.pdf.

[38] Shapley LS. A value for n-person games (1953). In: Roth AE, editor. The Shapley value: essays in honor of lloyd s. Shapley. Cambridge: Cambridge University Press; 1988, p. 31–40. http://dx.doi.org/10.1017/CBO9780511528446.003, URL https://www.cambridge.org/core/books/shapley-value/value-for-nperson-games/1AA9D343DE7A87A97F69E999D329B57A.

[39] Chen H, Janizek JD, Lundberg S, Lee S-I. True to the model or true to the data? 2020, URL http://arxiv.org/abs/2006.16234. arXiv:2006.16234 [cs, stat].

[40] Chen H, Covert IC, Lundberg SM, Lee S-I. Algorithms to estimate Shapley value feature attributions. Nat Mach Intell 2023;5(6):590–601. http://dx.doi.org/10.1038/s42256-023-00657-x, URL https://www.nature.com/articles/s42256-023-00657-x.

[41] Štrumbelj E, Kononenko I. Explaining prediction models and individual predictions with feature contributions. Knowl Inf Syst 2014;41(3):647–65. http://dx.doi.org/10.1007/s10115-013-0679-x, URL http://link.springer.com/10.1007/s10115-013-0679-x.

[42] Jedrzejewski A, Lago J, Marcjasz G, Weron R. Electricity price forecasting: The dawn of machine learning. IEEE Power Energy Mag 2022;20(3):24–31. http://dx.doi.org/10.1109/MPE.2022.3150809, URL https://ieeexplore.ieee.org/document/9761111/. Conference Name: IEEE Power and Energy Magazine.